\documentclass[lettersize,journal]{IEEEtran}
\usepackage{amsmath,amsfonts}
\usepackage{algorithmic}
\usepackage{algorithm}
\usepackage{array}
\usepackage[caption=false,font=normalsize,labelfont=sf,textfont=sf]{subfig}
\usepackage{textcomp}
\usepackage{stfloats}
\usepackage{url}
\usepackage{verbatim}
\usepackage{graphicx}
\usepackage{cite}
\usepackage{bm}
\usepackage{epstopdf}
\usepackage{multirow}
\usepackage{amsmath}
\usepackage{hyperref}

\newtheorem{definition}{\bf Definition}

\begin{document}

\title{Hierarchical Sparse Representation Clustering for High-Dimensional Data Streams}

\author{Jie~Chen, ~\IEEEmembership{Member,~IEEE,}
        Hua~Mao,
        Yuanbiao~Gou,
        and Xi~Peng, ~\IEEEmembership{Senior Member,~IEEE}
\thanks{Manuscript received \today. (Corresponding author: Xi~Peng.)}

\thanks{Jie Chen, Yuanbiao Gou and Xi Peng are with the College of Computer Science, Sichuan University,
Chengdu 610065, China (E-mail: chenjie2010@scu.edu.cn; gouyuanbiao@gmail.com; pengx.gm@gmail.com).}

\thanks{Hua Mao is with the Department of Computer and Information Sciences, Northumbria University, Newcastle, NE1 8ST, U.~K. (E-mail: hua.mao@northumbria.ac.uk).}}

\markboth{Journal of \LaTeX\ Class Files,~Vol.~14, No.~8, September~2024}%
{Shell \MakeLowercase{\textit{et al.}}: Hierarchical Sparse Representation Clustering For High-Dimensional Data Streams}

\IEEEpubid{0000--0000/00\$00.00~\copyright~2021 IEEE}

\maketitle

\begin{abstract}
Data stream clustering reveals patterns within continuously arriving, potentially unbounded data sequences. Numerous data stream algorithms have been proposed to cluster data streams. The existing data stream clustering algorithms still face significant challenges when addressing high-dimensional data streams. First, it is intractable to measure the similarities among high-dimensional data objects via Euclidean distances when constructing and merging microclusters. Second, these algorithms are highly sensitive to the noise contained in high-dimensional data streams. In this paper, we propose a hierarchical sparse representation clustering (HSRC) method for clustering high-dimensional data streams. HSRC first employs an $l_1$-minimization technique to learn an affinity matrix for data objects in individual landmark windows with fixed sizes, where the number of neighboring data objects is automatically selected. This approach ensures that highly correlated data samples within clusters are grouped together. Then, HSRC applies a spectral clustering technique to the affinity matrix to generate microclusters. These microclusters are subsequently merged into macroclusters based on their sparse similarity degrees (SSDs). Additionally, HSRC introduces sparsity residual values (SRVs) to adaptively select representative data objects from the current landmark window. These representatives serve as dictionary samples for the next landmark window. Finally, HSRC refines each macrocluster through fine-tuning. In particular, HSRC enables the detection of outliers in high-dimensional data streams via the associated SRVs. The experimental results obtained on several benchmark datasets demonstrate the effectiveness and robustness of HSRC.
\end{abstract}


\section{Introduction}
\label{sec:Introduction}
\IEEEPARstart{D}{ata} stream clustering techniques have attracted attention from data mining and machine learning researchers \cite{Gama2013SL, Silva2013DSC, Amini2014DSC, Nguyen2015DSC, Hahsler2016CDS}. They have significant impacts on the development of data stream applications \cite{Spiliopoulou2006MMCT}, such as network media transmission, sensor transfer information, and computer network traffic monitoring approaches. These applications involve data streams that can be potentially infinite, but only a limited amount of memory resources are available for computation purposes. Extracting potentially valuable knowledge from massive data streams naturally leads to the challenging problem of data stream clustering. As an effective tool for data stream mining, data stream clustering aims to divide the given data objects into a finite number of classes and identify potential outliers.

Traditional clustering algorithms, e.g., spectral clustering \cite{Shi2000Ncuts, Belkin2001LE}, statistical learning-based methods and their variants \cite{Huang2018MFA}, often achieve impressive performance on static and stable datasets. A data stream consists of massive, unbounded sequences of data objects, and its probability distribution changes over time in an unexpected way. This change can be gradual, which is known as concept drift, or sudden, which is referred to as concept shift. As a result, once entirely new classes appear in the stream, traditional classifiers are expensive and time-consuming, and prediction models need to be reconstructed and retrained from scratch via a newly collected data stream.

A variety of clustering methods have been proposed for data streams \cite{Guha2003CDS, Tu2009SDC, Amini2011DG, Shindler2011KM, Zhang2014DSC, Hyde2017EDS}. For example, an efficient and scalable data clustering algorithm, called balanced iterative reducing and clustering using hierarchies (BIRCH), was proposed for conducting exploratory analyses on very large datasets \cite{Zhang1997BIRCH}. BIRCH constructs a height-balanced tree via a clustering feature (CF) vector to compute cluster measures, such as the clustering mean, radius and diameter. The Euclidean distance between a new object and each centroid of the CF entries is calculated and later compared with a given threshold. The CF vector introduced by BIRCH has been extensively employed by different algorithms \cite{Aggarwal2003Framework, Feng2006DC, Kranen2011ClusTree}. Specifically, the concept of the CF vector was extended to microclusters in the CluStream algorithm \cite{Aggarwal2003Framework}, which includes two stages: online microclustering and offline macroclustering. The density-based data stream clustering (DenStream) algorithm \cite{Feng2006DC} introduces the structures of potential core microclusters and outlier microclusters via their CFs for incremental computation purposes. Additionally, the ClusTree algorithm uses a hierarchical indexing structure to maintain CFs and further builds a microcluster hierarchy at different granularity levels \cite{Kranen2011ClusTree}.

Data stream clustering algorithms utilize CF vectors to measure the similarity among data objects, which is intrinsically tied to the construction and merging operations involving appropriate microclusters. However, these approaches still face significant challenges.
\IEEEpubidadjcol
First, a new data object is merged into its nearest microcluster if the new radius (determined based on the associated CFs) is smaller than a predetermined threshold. Selecting an appropriate maximum boundary for this radius is a challenging problem. Second, these algorithms often rely on local similarity measures, typically the Euclidean distances between data objects, to maintain local neighborhood information. However, noise is pervasive in high-dimensional data streams. The use of the Euclidean distance measure for evaluating the relationships among data objects makes the similarity computation highly sensitive to noise. Clustering inherently assumes that highly correlated data objects within the same cluster share similar structural characteristics. Therefore, measuring the similarity of the data objects contained in high-dimensional data streams remains a significant challenge.

Recently, sparse representation has received significant interest across various fields \cite{Chen2022TSSRC, Chen2022OSRC, Olshausen1997SC, Tibshiran1996Lasso, Donoho2006L1}, including image classification \cite{Chen2024SEF, Chen2018SLLP}, image denoising \cite{Elad2006ImageDenoising}, and visual tracking \cite{Yuan2012MultiSR}. It offers a statistical model for identifying a set of dictionary elements from the same class for each data object. Ideally, sparse representation groups highly correlated data objects belonging to the same class, with the nonzero elements representing the weights assigned to different pairs of data objects. As a result, sparse representation has demonstrated the ability to mitigate the aforementioned challenges by leveraging the self-similarity properties of data objects contained in high-dimensional data streams.

In this paper, we propose a hierarchical sparse representation clustering (HSRC) method for clustering high-dimensional data streams. HSRC first incorporates a sparse constraint into the self-expressiveness property of data objects in high-dimensional data streams. It aims to learn an affinity matrix that can capture the relationships among different data objects. For example, given a set of data objects in a landmark window, each object is represented as a linear combination of the other objects, and the coefficient matrix is sparse. In particular, we integrate adaptive dictionary learning into the sparse representation process to improve its generalization in data stream clustering scenarios. Then, HSRC partitions the affinity matrix into microclusters via a spectral clustering technique. The sparse linear representation of each data object includes more objects from the same cluster, indicating that the affinity matrix encodes the relationships within microclusters. Unlike density-based clustering techniques, HSRC does not rely on Euclidean distances to measure relationships among data objects. Instead, two microclusters are merged into a new macrocluster if the sparse similarity degree (SSD) of the merged cluster is higher than that between the new microcluster and any other microcluster. Finally, the clustering labels of the data objects contained within the macroclusters are fine-tuned to enhance the performance of the data stream clustering process. Additionally, HSRC includes an outlier detection mechanism for identifying potential outliers.

The contributions of this paper can be summarized as follows.
\begin{enumerate}
\item We present an HSRC model that measures the relationships among data objects via sparse representation.
\item A residual sparsity value is introduced for conducting adaptive dictionary sample selection and outlier identification, which can improve the generalization of the sparse representation process in high-dimensional data stream clustering tasks.
\item The SSDs among microclusters are introduced to construct macroclusters, which together form a hierarchical structure for performing high-dimensional data stream clustering.
\item Extensive experimental results obtained on benchmark datasets demonstrate the effectiveness and robustness of HSRC in high-dimensional data stream clustering scenarios.
\end{enumerate}

The remainder of this paper is organized as follows. The related work on sparse representation and data stream clustering is summarized in Section \ref{sec:Relatedwork}. Section \ref{sec:HSRC} describes the HSRC method in detail. Extensive experimental results obtained several benchmark datasets are presented in Section \ref{sec:Exp}. Finally, conclusions are given in Section \ref{sec:Conclusion}.

\section{Related Work}
\label{sec:Relatedwork}
In this section, we present a brief overview of the related work on sparse representation and data stream clustering.

\subsection{Sparse Representation Theory}
Let $\mathbf{D} = [{\mathbf{d}_1},{\mathbf{d}_2},...,{\mathbf{d}_n}] \in {\mathbf{R}^{d \times n}}$ be a dictionary consisting of $n$ vectors. Given a signal $x = \in {\mathbf{R}^{d}}$, its sparsest representation over the dictionary $\mathbf{D}$ can be approximately represented by a sparse linear combination of only a few columns of $\mathbf{D}$, i.e.,
\begin{equation}\label{eq:sr1}
\mathop {\min }\limits_{ \mathbf{z}} {\left\| {\mathbf{x} - \mathbf{D}\mathbf{z} } \right\|_2} + \lambda {\left\| \mathbf{z}  \right\|_0},
\end{equation}
where $\mathbf{z}$ represents a sparse coefficient, $\lambda > 0$ is a tradeoff parameter, and ${\left\|  \cdot  \right\|_0}$ denotes the number of nonzero elements in a vector.

Since Problem \eqref{eq:sr1} is nonconvex and nondeterministic polynomial-time (NP)-hard, it is often relaxed to the following ${l_1}$-norm minimization problem as a common surrogate for the sparse representation task, i.e.,
\begin{equation}\label{eq:sr2}
\mathop {\min }\limits_{ \mathbf{z}}  {\left\| {\mathbf{x} -  \mathbf{D}\mathbf{z} } \right\|_2} + \lambda {\left\| { \mathbf{z}}  \right\|_1},
\end{equation}
where ${\left\|  \cdot  \right\|_1}$ denotes the $l_1$-norm of a vector. The above optimization problem can be solved via various convex optimization methods, such as iterative thresholding algorithms \cite{Tibshirani1996RS, Donoho2006L1}.

Inspired by the advances achieved with respect to ${l_0}$-norm and ${l_1}$-norm techniques, sparse representation has been widely used in various areas of machine learning and pattern recognition \cite{Natarajan1995NPHard, Bruckstein2009SparseSolutions}. For example, Aharon et al. \cite{Aharon2006KSVD} proposed a $K$-singular value decomposition (SVD) algorithm for designing overcomplete dictionaries for sparse representation purposes. The $K$-SVD algorithm can effectively seek the best representation of a given set as a dictionary under strict sparsity constraints. Hence, sparse representation has been proven to be an extremely successful technique for exploiting the intrinsic structures of high-dimensional data.

\subsection{Data Stream Clustering Techniques}
A data stream is a massive, potentially unbounded sequence of data objects which are continuously generated over time. Most existing data stream clustering algorithms can be broadly categorized into three main types: hierarchy-based clustering, density-based clustering and partitioning-based clustering.

Hierarchy-based clustering algorithms first use the data objects to construct a tree, in which each leaf represents a data object. These algorithms then group data objects into the corresponding clusters using agglomerative or divisive hierarchical decomposition. In the agglomerative approach, $n$ objects are merged into more general classes in a bottom-up fashion, whereas in divisive techniques, $n$ objects are divided into smaller clusters in a bottom-up fashion. BIRCH is a typical hierarchical clustering algorithm that constructs a height-balanced tree using the CF vector \cite{Zhang1997BIRCH}. BIRCH adopts the CF vector to compute cluster measures such as the mean, radius, and diameter. Finally, the Euclidean distances between new objects and each centroid of the CF entries are calculated and compared against a threshold to determine whether the new object belongs to a new class. The BIRCH algorithm has $O(n)$ complexity and obtains good clustering results through only a single scan. However, it does not work well with data of arbitrary shape. The E-Stream algorithm is an evolution-based approach for data stream clustering that supports five evolutions of the data, namely appearance, disappearance, self-evolution, merging, and splitting \cite{Udommanetanakit2007ES, L2009IC}.

Density-based clustering algorithms define clusters as high-density areas of the features, which can be used to detect any arbitrary-shaped clusters. DenStream, an extended version of the DBSCAN algorithm, discovers clusters of arbitrary shape in an evolving data stream \cite{Feng2006DC}. It introduces a core microcluster, a potential core microcluster, and an outlier microcluster using the CFs developed for BIRCH. The core microcluster is adopted to summarize the clusters with an arbitrary shape, while the potential core microcluster and outlier microcluster structures are used to pursue and distinguish potential clusters and outliers. DenStream involves numerous time vector calculations in the offline clustering phase, which leads to a high computational cost. The D-Stream algorithm automatically and dynamically adjusts the clusters without requiring user specification of the target time horizon or the number of clusters \cite{Tu2009SDC}. An online component maps each input data record into a grid, while an offline component computes the grid density and clusters the grids based on the density. However, D-Stream is incapable of processing very-high-dimensional data.

Partitioning-based clustering methods use traditional soft partitioning techniques such as $k$-median and $k$-means clustering to deal with data streams. CluStream is a two-phase clustering method that performs online micro-clustering and offline macro-clustering \cite{Aggarwal2003Framework}. The first phase involves the acquisition of summary statistics from the data stream by extending the concept of the CF vector in the online micro-clustering. Each microcluster contains five components, three of which are the regular components of the CF vector. The second phase uses these statistics and other inputs to create clusters, where the $k$-means algorithm is embedded in the offline macro-clustering component. HPStream was developed as an extension to CluStream that performs projected clustering for high-dimensional streaming data. CluStream is not efficient when applied to high-dimensional data streams.

\section{Hierarchical Sparse Representation Clustering}
\label{sec:HSRC}
In this section, we introduce the HSRC method for clustering high-dimensional data streams. Given the potentially infinite nature of data streams, HSRC employs landmark windows, thereby sequentially processing fixed-size, nonoverlapping chunks of data objects. The proposed HSRC model contains three critical processes, including the creation of microclusters, the merging of microclusters into macroclusters and the fine-tuning of macroclusters, which together form a hierarchical structure. Throughout these processes, the data object representatives are repeatedly utilized across three successive HSRC processes. Therefore, the proposed HSRC method can effectively capture the intrinsic structures of the high-dimensional data objects contained within data streams.

\subsection{Evaluating the Relationships Among Data Objects via Sparse Representation}
Let $\mathbf{X} = [{\mathbf{x}_1},{\mathbf{x}_2},...,{\mathbf{x}_n}] \in {\mathbf{R}^{d \times n}}$ be a set of data objects included in a landmark window. The rationale assumption behind clustering algorithms is that data objects within a cluster are more similar to each other than they are to objects belonging to a different cluster. Given a data object ${\mathbf{x}_i}$ ($1 \le i \le n$), HSRC uses the following ${l_1}$-norm minimization problem for sparse representation:
\begin{equation}\label{eq:sr3}
\mathop {arg\min }\limits_{\mathbf{z}_i} {\left\| {\mathbf{z}_i} \right\|_1} \qquad s.t.  \qquad {\left\| {{\mathbf{x}_i} - \mathbf{X}{\mathbf{z}_i}} \right\|_2^2} \le \varepsilon
\end{equation}
where $\varepsilon$ represents a noise item. The above optimization problem can be solved in polynomial time via standard linear programming algorithms \cite{Tibshirani1996RS, Donoho2006L1}. Here $\mathbf{Z}$ can be constructed using the vectors of sparse coefficients, i.e., $\mathbf{Z} = [{\mathbf{z}_1},{\mathbf{z}_2},...,{\mathbf{z}_n}]$, and ${\mathbf{Z}^*} = \left| \mathbf{Z} \right| + \left| {{\mathbf{Z}^T}} \right|$. Each element $z_{ij}$ in ${\mathbf{Z}^*}$ can be used to measure the similarity between two data objects ${\mathbf{x}_i}$ and ${\mathbf{x}_j}$. If the data objects $\mathbf{x}_i$ and $\mathbf{x}_j$ are close in terms of the intrinsic geometries of their data distributions, their representations, $\mathbf{z}_i$ and $\mathbf{z}_j$, respectively, should also be close with respect to the same basis $\mathbf{X}$. Hence, sparse representation is utilized to evaluate the relationships among the data objects contained in high-dimensional data streams.

The solution to Problem \eqref{eq:sr3} is individually obtained for each data object. This inevitably results in high computational costs. However, clustering high-dimensional data streams requires data objects to be continuously clustered within a limited period. To increase the efficiency of optimizing individual sparse representations, we use a sparse batch representation optimization problem as a good surrogate for Problem (3). Specifically, we formulate the following convex optimization problem to find a sparse representation $\mathbf{Z}$:
\begin{equation}\label{eq:sr4}
\min \limits_{\mathbf{Z}, \mathbf{E}} {\left\| \mathbf{Z} \right\|_1} + \lambda \left\| {\mathbf{E}} \right\|_l \qquad s.t. \qquad  \mathbf{X} = \mathbf{X}\mathbf{Z} + \mathbf{E},
\end{equation}
where $\mathbf{E} \in {\mathbf{R}^{d \times n}}$ is a noise item and $\lambda$ is a scalar constant.

\begin{algorithm}
\renewcommand{\algorithmicrequire}{\textbf{Input:}}
\renewcommand\algorithmicensure {\textbf{Output:} }
\caption{Solving Problem \eqref{eq:sr5} via an inexact ALM framework }
\label{alg:HSRC1}
\begin{algorithmic}
\REQUIRE ~~\\
a data matrix $\mathbf{X} = [{\mathbf{x}_1},{\mathbf{x}_2},...,{\mathbf{x}_n}] \in {\mathbf{R}^{d \times n}}$, a parameter $\lambda > 0$
\end{algorithmic}
{\bfseries Initialize:}
$\mathbf{Z}=\mathbf{J}=0, \mathbf{E}=0, {\mathbf{Y}_1}={\mathbf{Y}_2}=0, {\mu}={10^{-2}}, $ \\
${\mu_{\max }}={10^{10}},\rho=1.1,\varepsilon = {10^{ - 6}}$ \\
\begin{algorithmic}[1]
\WHILE {not converged}
\STATE update the variables via Eq. \eqref{eq:sr8}; \\
\STATE update the multipliers: \\
${\mathbf{Y}_1} \leftarrow {\mathbf{Y}_1} + {\mu} \left(\mathbf{X} - \mathbf{X}\mathbf{Z} - \mathbf{E}\right)$; \\${\mathbf{Y}_2}  \leftarrow  {\mathbf{Y}_2} + {\mu} \left(\mathbf{Z} - \mathbf{J}\right)$;
\STATE update the parameter ${\mu}$: \\
${\mu} \leftarrow \min (\rho {\mu},{\mu _{\max}})$;\\
\STATE check the convergence conditions: \\
\begin{center}
${\left\| {\mathbf{X} - \mathbf{X}\mathbf{Z} - \mathbf{E}} \right\|_{max}} < \varepsilon $ and ${\left\| {\mathbf{Z} - \mathbf{J}} \right\|_{max}} < \varepsilon $;
\end{center}
\ENDWHILE
\ENSURE ~~\\ $ \left( {\mathbf{Z}, \mathbf{E}} \right)$
\end{algorithmic}
\end{algorithm}

We first convert Problem \eqref{eq:sr4} to the following equivalent problem by introducing an auxiliary variable $\mathbf{J}$:
\begin{equation}\label{eq:sr5}
\min \limits_{\mathbf{Z}, \mathbf{J}, \mathbf{E}} {\left\| \mathbf{J} \right\|_1} + \lambda \left\| {\mathbf{E}} \right\|_l \qquad s.t. \qquad  \mathbf{X} = \mathbf{X}\mathbf{Z} + \mathbf{E}, \ \mathbf{Z}=\mathbf{J}.
\end{equation}

The augmented Lagrangian function of Problem \eqref{eq:sr5} is
\begin{equation}\label{eq:sr6}
\begin{split}
& \mathop {\min }\limits_{\mathbf{Z},\mathbf{E},\mathbf{J},{\mathbf{Y}_1},{\mathbf{Y}_2}} {\left\| \mathbf{J} \right\|_1} + \lambda \left\| \mathbf{E} \right\|_l + tr\left( {\mathbf{Y}_1^T(\mathbf{X} - \mathbf{X}\mathbf{Z} - \mathbf{E}} \right) + \\
& tr\left( {\mathbf{Y}_2^T\left( {\mathbf{Z} - \mathbf{J}} \right)} \right) + \frac{{{\mu }}}{2}\left(\left\| {\mathbf{X} - \mathbf{X}\mathbf{Z} - \mathbf{E}} \right\|_F^2 + \left\| {\mathbf{Z} - \mathbf{J}} \right\|_F^2\right),
\end{split}
\end{equation}
where ${\mathbf{Y}_1}$ and ${\mathbf{Y}_2}$ are Lagrange multipliers, and ${\mu_1} > 0$ and ${\mu_2} > 0$ are penalty parameters. The above optimization problem can be formulated as follows:
\begin{equation}\label{eq:sr7}
\begin{split}
& \mathop {\min }\limits_{\mathbf{Z},\mathbf{E},\mathbf{J},{\mathbf{Y}_1},{\mathbf{Y}_2}} {\left\| \mathbf{J} \right\|_1} + \lambda \left\| \mathbf{E} \right\|_l + \\
& \frac{{{\mu}}}{2}\left(\left\| {\mathbf{X} - \mathbf{X}\mathbf{Z} - \mathbf{E} + \frac{{{\mathbf{Y}_1}}}{{{\mu}}}} \right\|_F^2 + \left\| {\mathbf{Z} - \mathbf{J} + \frac{{{\mathbf{Y}_2}}}{{{\mu}}}} \right\|_F^2\right).
\end{split}
\end{equation}
This formula can be effectively solved by the inexact augmented Lagrange multiplier (ALM) framework \cite{Lin2011LADM}. The variables $\mathbf{J}$, $\mathbf{Z}$ and $\mathbf{E}$ can be updated alternately at each step, whereas the other two variables are fixed. The updating schemes for the $(k+1)$ th iteration are as follows:
\begin{equation}\label{eq:sr8}
\begin{split}
& {\mathbf{J}} \leftarrow \mathop {\min }\limits_{\mathbf{J}} \frac{1}{{{\mu}}}{\left\| \mathbf{J} \right\|_1} + \frac{1}{2}\left\| {\mathbf{J} - \left( {\mathbf{Z} + \frac{{{\mathbf{Y}_2}}}{{{\mu}}}} \right)} \right\|_F^2, \\
& {\mathbf{Z}} \leftarrow \left( {\mathbf{I} + {\mathbf{X}^T}\mathbf{X}} \right)^{ - 1}\left( {{\mathbf{X}^T}\mathbf{X} - {\mathbf{X}^T}\mathbf{E} + \mathbf{J} + \frac{{{\mathbf{X}^T}{\mathbf{Y}_1} - {\mathbf{Y}_2}}}{{{\mu}}}} \right), \\
& {\mathbf{E}} \leftarrow \mathop {\min }\limits_{\mathbf{E}} \lambda {\left\| \mathbf{E} \right\|_l} + \frac{{{\mu}}}{2}\left\| {\mathbf{X} - \mathbf{X}\mathbf{Z} - \mathbf{E} + \frac{{{\mathbf{Y}_1}}}{{{\mu}}}} \right\|_F^2.
\end{split}
\end{equation}

\begin{algorithm}[!htbp]
\renewcommand{\algorithmicrequire}{\textbf{Input:}}
\renewcommand\algorithmicensure {\textbf{Output:} }
\caption{Creating microclusters}
\label{alg:HSRC2}
\begin{algorithmic}[1]
\REQUIRE ~~\\
a data matrix $\mathbf{X} = [{\mathbf{x}_1},{\mathbf{x}_2},...,{\mathbf{x}_n}] \in {\mathbf{R}^{d \times n}}$; parameters $\lambda > 0$, $m > 0$ and $\sigma  \in \left( {0,1} \right)$
\STATE Solve Problem \eqref{eq:sr5} via Algorithm \ref{alg:HSRC1} and obtain the optimal solution $ \left( {\mathbf{Z}, \mathbf{E}} \right)$.
\STATE Compute the affinity matrix ${\mathbf{Z}^*} = \left| \mathbf{Z} \right| + \left| {{\mathbf{Z}^T}} \right|$.
\STATE Apply NCuts on ${\mathbf{Z}^*}$ to obtain $m$ microclusters.
\STATE For a new data object $\mathbf{x}$, compute $SRV(\mathbf{x})$ using Eq. \eqref{eq:srv};
\IF{ $SRV(\mathbf{x}) < \sigma$ }
\STATE $\mathbf{x}$ is considered an outlier.
\ENDIF
\ENSURE ~~\\ The $m$ microclusters.
\end{algorithmic}
\end{algorithm}

The first equation in Problem \eqref{eq:sr8} is a convex problem that can be solved via the singular value thresholding (SVT) operator \cite{Cai2010SVT}. The solution of the third equation in Problem \eqref{eq:sr8} is closely related to the value of the $l$-norm, which indicates a certain regularization strategy for characterizing various noises. For example, the ${l_{2,1}}$-norm encourages the columns of the matrix to be zero, where the last equation has a closed-form solution \cite{Liu2010LRR}. The complete procedure for solving Problem \eqref{eq:sr5} is outlined in Algorithm \ref{alg:HSRC1}.

\subsection{Evaluating the Importance of Data Objects via Sparsity Residual Values}
HSRC uses the learned affinity matrix to identify clusters as groups of microclusters through a spectral clustering algorithm, such as NCuts \cite{Shi2000Ncuts}. For simplicity, we assume that $n$ data objects in a landmark window are segmented into $m$  microclusters, i.e., $\mathbf{X} = [{\mathbf{X}_1},{\mathbf{X}_2},...,{\mathbf{X}_m}]$, according to Algorithm \ref{alg:HSRC2}. For example, the $i$th microcluster contains $n_i$ data objects, i.e., ${\mathbf{X}_i} = [{\mathbf{X}_{i1}},{\mathbf{X}_{i2}},...,{\mathbf{X}_{i{n_i}}}]$. Each microcluster consists of a set of neighboring objects, where each data object within the microcluster can be linearly represented by other objects in the cluster. The microclusters are identified in a single pass of the landmark window, and their summary statistics are stored offline.

\begin{definition}
Sparsity residual value (SRV): The SRV of a data object $\mathbf{x}$ is defined as
\begin{equation}\label{eq:srv}
SRV(\mathbf{x}) = \frac{1}{{{{\left\| \mathbf{e} \right\|}_0}}}\sum\limits_{j = 1}^{n_i} {\frac{{\left| {{e_j}} \right|}}{{{{\left\| \mathbf{e} \right\|}_2}}}}
\end{equation}
where $\mathbf{e} \in {\mathbf{R}^{n_i}}$ denotes a corresponding noise item in the sparse representation of the object and $n_i$ is the number of samples contained in the $i$th cluster.
\end{definition}

We define SRVs to measure the importance levels of the data objects included in clusters. The lower the SRV value of a data object is, the more significant it is during the sparse representation process. Without loss of generality, the sparse coefficients of a data object $\mathbf{x}$ are represented by a vector $\mathbf{z}$, and $\mathbf{e} $ denotes a corresponding noise item in the sparse representation of the object. Given a solution $\mathbf{Z}$ for the data objects $\mathbf{X}$ obtained by Algorithm \ref{alg:HSRC1}, the representative data objects can be adaptively selected from the candidates by sorting their SRVs. The sizes of the representative data objects are aligned with those of the data objects contained in the landmark window. The dictionary samples in an adaptive dictionary selection scheme consist of two parts, including the data objects contained in the current landmark window and the representative data objects chosen from the previous landmark window.

\begin{algorithm}[!htbp]
\renewcommand{\algorithmicrequire}{\textbf{Input:}}
\renewcommand\algorithmicensure {\textbf{Output:} }
\caption{Merging microclusters into macroclusters}
\label{alg:HSRC3}
\begin{algorithmic}
\REQUIRE ~~\\
Microclusters $\mathbf{X} = [{\mathbf{X}_1},{\mathbf{X}_2},...,{\mathbf{X}_m}]$ and an affinity matrix ${\mathbf{Z}^*}$
\end{algorithmic}
{\bfseries Initialize:}
$\mathbf{S} \in {\mathbf{R}^{m \times m}} = 0$;
\begin{algorithmic}[1]
\WHILE {the clusters in $\mathbf{X}$ have been changed }
\FOR{each cluster $i$ among the clusters}
\STATE $s$ $ \leftarrow $ the number of clusters in $\mathbf{X}$;
\FOR{each cluster $j$ among the clusters and $i \ne j$}
\STATE Compute $S({\mathbf{X}_i},{\mathbf{X}_j})$ and $S({\mathbf{X}_{i+j}},{\mathbf{X}_p})$ using Eqs. \eqref{eq:mc1} and \eqref{eq:mc2}, respectively, where $p \in \left[ {1,s} \right]$, $p \ne i$ and $p \ne j$; \\
\IF{ $ \mathbf{S}({\mathbf{X}_i},{\mathbf{X}_j}) \le \mathbf{S}({\mathbf{X}_{i+j}},{\mathbf{X}_p}) $ }
\STATE $\mathbf{S}[ij]=1$;
\ENDIF
\ENDFOR
\STATE $mergedSet = []$,
\FOR { $i = 1 : s$ and $i \notin mergedSet$}
\STATE $clusterRows = find(S(i, :) == 1)$;
\STATE $clusterCols = find(S(:, i) == 1)$;
\STATE $clusterSet = clusterRows \cap clusterCols$
\IF{ $ \sim isempty(clusterSet)$ }
\STATE Merge cluster $i$ and all clusters in $clusterSet$;
\STATE Add $i$ and all items of $clusterSet$ to $mergedSet$;
\ENDIF
\ENDFOR
\ENDFOR
\ENDWHILE
\ENSURE ~~\\ The clusters in $\mathbf{X}$ .
\end{algorithmic}
\end{algorithm}

In practice, outliers often arise due to various factors, such as data collection, storage, or transmission errors. Outliers are data objects that deviate from a sparse representation-based model. To identify outliers throughout high-dimensional data streams, we set a threshold $\sigma  \in \left( {0,1} \right)$. A data object is estimated according to the threshold $\sigma$. For example, we consider a data object an outlier if it satisfies the following condition:
\begin{equation}\label{eq:condtion}
SRV(\mathbf{x}) \ge  \sigma.
\end{equation}
HSRC is able to adaptively identify outliers in high-dimensional data streams. Algorithm \ref{alg:HSRC2} summarizes the complete microclustering algorithm of HSRC.

\begin{algorithm}[!htbp]
\renewcommand{\algorithmicrequire}{\textbf{Input:}}
\renewcommand\algorithmicensure {\textbf{Output:} }
\caption{Fine-tuning macroclusters}
\label{alg:HSRC4}
\begin{algorithmic}
\REQUIRE ~~\\
$s$ macroclusters $\mathbf{X} = [{\mathbf{X}_1},{\mathbf{X}_2},...,{\mathbf{X}_s}] \in {\mathbf{R}^{d \times n}}$, a parameter $\sigma  \in \left( {0,1} \right)$
\end{algorithmic}
\begin{algorithmic}[1]
\FOR{each cluster $i$ in the macroclusters}
\FOR{each data object $\mathbf{x}$ in the $i$th cluster}
\FOR{each cluster $j$ in the macroclusters}
\STATE Construct an error set using Eq. \eqref{eq:ft1} as follows: \\
$errorSet = [error_{1}, {error_{2}}, ..., {error_{s}}]$;
\STATE Determine the clustering index $j$ of the minimum value in the error set;
\IF{ $ i  \ne j $ }
\STATE Select $\mathbf{x}$  from the $i$th macrocluster and drop it into the $j$th macrocluster;
\ENDIF
\ENDFOR
\ENDFOR
\ENDFOR
\ENSURE ~~\\ The final clusters.
\end{algorithmic}
\end{algorithm}

\subsection{Merging Micro-clusters into Macro-clusters}
As microclusters are rough, they need to be merged into macroclusters. Considering the $i$th microcluster, we first choose a candidate from another microcluster as the merged target. Utilizing only the sparse coefficients associated with the $j$th microcluster ${\mathbf{X}_j}$, the sparse representation of the $l$th data object ${\mathbf{x}_{l}}$ contained in the $i$ th microcluster can be written as follows:
\begin{equation}\label{eq:mc1}
{\mathbf{x}_{l}} = {\mathbf{X}_j}{\mathbf{z}_{j}} + {\mathbf{e}_{l}},
\end{equation}
where $i,j \in [1,m]$, $l \in [1,n_i]$ and $i \ne j$. The item $\mathbf{z}_{j}  \in {\mathbf{R}^{n_j}}$ is a sparse coefficient vector whose entries are those associated with $\mathbf{X}_j$. The residual item ${\mathbf{e}_{l}}$ can be used to measure the similarity between ${\mathbf{x}_{l}}$ and ${\mathbf{X}_j}$. We adopt the residual item to define the SSD concept, which is used to measure the similarity between two microclusters in the microcluster merging operation.

\begin{definition}
SSD: The SSD between cluster $i$ and cluster $j$ is the sum of the residuals of the data objects contained in the $i$th microcluster ${\mathbf{X}_i}$ that are associated with the $j$th microcluster ${\mathbf{X}_j}$, which can be defined as
\begin{equation}\label{eq:mc2}
S({\mathbf{X}_i},{\mathbf{X}_j})= \sum\limits_{l = 1}^{n_i} {{\left\| {\mathbf{e}_{l}} \right\|}_2},
\end{equation}
where $i \ne j$ and $i,j \in [1,m]$.
\end{definition}

We further calculate the SSDs of cluster $i$ with the other clusters. Finally, we can obtain a candidate based on these residual items by choosing the lowest error. We need to further check the merging condition to determine whether cluster $i$ and cluster $j$ should be merged. We assume that we obtain a new cluster ${\mathbf{X}_{i+j}}$ after clusters $i$ and $j$ are merged. According to graph theory, the merging condition is that the similarity degree of the merged cluster ${\mathbf{X}_{i+j}}$ must be greater than that of cluster ${\mathbf{X}_{i+j}}$ associated with any other clusters ${\mathbf{X}_{p}}$; i.e., the following two conditions must hold:
\begin{equation}
\begin{split}
& S({\mathbf{X}_i},{X_j}) \le S({\mathbf{X}_{i+j}},{\mathbf{X}_p}), \\
& S({\mathbf{X}_j},{X_i}) \le S({\mathbf{X}_{i+j}},{\mathbf{X}_p}),
\end{split}
\end{equation}
where $p \ne i$ and $p \ne j$. If the above conditions hold, two microclusters are merged into a relatively large cluster. The final merging operation is stopped if any two microclusters cannot be further merged. For simplicity, we define a new matrix $\mathbf{S}$ with a size of $m\times m$ to preserve the merged sign for each pair of $m$ microclusters. For instance, ${S_{ij}}$ denotes that cluster $i$ can be merged into cluster $j$. Algorithm \ref{alg:HSRC3} below summarizes the complete microclusters merging procedure.

\begin{algorithm}[!htbp]
\renewcommand{\algorithmicrequire}{\textbf{Input:}}
\renewcommand\algorithmicensure {\textbf{Output:} }
\caption{The HSRC algorithm}
\label{alg:HSRC}
\begin{algorithmic}[1]
\REQUIRE ~~\\
$\mathbf{X}_t = [{\mathbf{x}_1},{\mathbf{x}_2},...,{\mathbf{x}_n}] \in {\mathbf{R}^{d \times n}}$, parameters $\lambda > 0$, $m > 0$ and $\sigma  \in \left( {0,1} \right)$
\IF{ $ t == 1 $ }
\STATE $\mathbf{X} = {\mathbf{X}_1}$ and $\mathbf{X}_s = \mathbf{X}_1$;
\ELSE
\STATE $\mathbf{X} = \left[ {{\mathbf{X}_s},{\mathbf{X}_{t}}} \right]$;
\ENDIF
\STATE Solve Problem \eqref{eq:sr5} via Algorithm \ref{alg:HSRC1} and obtain the optimal solution $ \left( {\mathbf{Z}, \mathbf{E}} \right)$.
\STATE Create $m$ microclusters via Algorithm \ref{alg:HSRC2}, and generate $m$ microclusters.
\STATE Perform the merging operation on the $m$ microclusters via Algorithm \ref{alg:HSRC3}, and obtain $s$ macroclusters.
\STATE Purify each of the $s$ macroclusters in a fine-tuning manner via Algorithm \ref{alg:HSRC4}, and obtain the final clusters.
\FOR{each cluster in the final clusters}
\FOR{each data object $\mathbf{x}$ in a cluster}
\STATE Computing $SRV\left( \mathbf{x} \right)$  using Eq. \eqref{eq:srv};
\IF{ $SRV(\mathbf{x}) \ge \sigma$ }
\STATE $\mathbf{x}$ is regarded as an outlier;
\ENDIF
\ENDFOR
\STATE Select the representative data objects from the cluster by sorting their SRVs and then add them to ${\mathbf{X}_s}$;
\ENDFOR
\ENSURE ~~\\ The final clusters.
\end{algorithmic}
\end{algorithm}

\subsection{Fine-Tuning}
Although sparse representation attempts to automatically choose data objects belonging to the same class in data streams, the coefficients of sparse representation are not completely concentrated on a particular microcluster and instead are likely to be widely spread across a few microclusters. Hence, the microclusters can be considered rough clusters. A data object has a sparse representation whose nonzero entries are concentrated mostly in one microcluster. The distribution of the estimated sparse coefficients contains important clustering information about the relationships among the data objects. Under these circumstances, fine-tuning is a strategy in which a few data objects are selected from a microcluster and then dropped into another microcluster under certain conditions. This strategy incrementally improves the clustering performance achieved for data objects.

From a sparse representation perspective, fine-tuning considers each data object in a microcluster with respect to another microcluster as follows:
\begin{equation}\label{eq:ft1}
{error_{j}} = \left\| {{\mathbf{x}_{l}} - {\mathbf{X}_j}{\mathbf{Z}_j}} \right\|_F^2,
\end{equation}
where $\mathbf{x}_{l}$ denotes the $l$th data object in the $i$th microcluster, and ${\mathbf{X}_j}$ is the $j$th microcluster. We assume that $s$ macroclusters are obtained via Algorithm  \ref{alg:HSRC3}. The error set of $\mathbf{x}_{l}$ is represented as $errorSet = [error_{1}, {error_{2}}, ..., {error_{s}}]$, and the clustering index of the minimum value included in the error set is $j$.
\begin{table}[!htbp]
\small
\setlength{\abovecaptionskip}{0pt}
\setlength{\belowcaptionskip}{0pt}
\setlength{\tabcolsep}{3pt}
\centering
\caption{ Descriptions of the streaming datasets.}
\label{tb:datasets}
\begin{tabular}{cccc}
\hline
Datasets & Classes & Data objects & Features \\
\hline
Keystroke & 4 & 1,600 & 10  \\
Network Intrusion & 2 & 494,000 & 42 \\
Forest Cover & 7 & 580,000 & 54 \\
COIL-100 & 100 & 7,200 & 1,024 \\
\hline
\end{tabular}
\end{table}
If $i \ne j$, $\mathbf{x}_{l}$ is selected from the $i$th microcluster and dropped into the $j$th microcluster. For example, a data object is picked up from its original cluster and dropped into a new microcluster if it has a smaller residual with respect to another microcluster.

Finally, it is critical to choose candidates from among the data objects that will be purified in their clusters. We consider these data objects, which indicate the spread of the sparse coefficients over most classes. In particular, we adopt the SRV as a criterion to help choose potential outliers in each cluster. Algorithm \ref{alg:HSRC4} below summarizes the complete macrocluster fine-tuning procedure.

\subsection{Computational Complexity Analysis}
We assume that $\mathbf{X}$ in a sliding-window has $n$ data objects that belong to $s$ classes, where the size of $\mathbf{X}$ is $d \times n$. Algorithm \ref{alg:HSRC} summarizes the complete data stream clustering algorithm of HSRC. We use an inexact ALM framework in Algorithm \ref{alg:HSRC1} \cite{Lin2011LADM}. The inexact ALM framework has been extensively studied and generally converges well. Algorithm \ref{alg:HSRC1} performs well in practical applications. The computational complexity of the first step of Algorithm \ref{alg:HSRC1} is $O({n^2})$ because it requires the sparse representation of an $n \times n$ matrix  to be computed in an SVT operator. The overall computational complexity of Algorithm \ref{alg:HSRC1} is $O(t{n^2})$ if the ${l_{2,1}}$-norm is adopted in the last equation of Problem \eqref{eq:sr8}, where $t$ is the number of iterations. When $n > d$, the computational complexity of Algorithm \ref{alg:HSRC2} can be considered $O({n^3})$ in spectral clustering. The computational complexities of Algorithm \ref{alg:HSRC3} and Algorithm \ref{alg:HSRC4} are $O({k^2}{n^2})$ and $O({s^2}{n^2})$ respectively, where $k$ and $s$ are the numbers of microclusters and macroclusters, respectively. The complexity of HSRC is $O({t{n^2} + {n^3} + {k^2}{n^2} + {s^2}{n^2}})$. Therefore, the final overall complexity of HSRC is $O({n^3})$ if $t \ll n$, $k \ll n$ and $s \ll n$.

\section{Experimental Study}
\label{sec:Exp}
In this section, we evaluate the performance of the proposed HSRC approach on publicly available datasets by comparing it with existing popular data stream clustering algorithms: CluStream \cite{Aggarwal2003Framework}, ClusTree \cite{Kranen2011ClusTree}, CluStreamKM \cite{ABifet2010MA}, StreamKM++ \cite{Ackermann2012SKM}, fuzzy double $c$-means based on sparse self-representation (FDCM\_SSR) \cite{Gu2018FDCC} and flexible density peak clustering (FDPC) \cite{Hou2024FDPC}. HSRC is implemented in MATLAB, and all the experiments are conducted on a Windows platform with an Intel i5-2300 CPU and 16 GB of RAM. The anonymous source code\footnote{https://github.com/chenjie20/HSRC} for SSCDL is available online. The implementations of the CluStream, ClusTree, CluStreamKM and StreamKM++ algorithms are provided by Massive Online Analysis (MOA), which is a popular open-source tool for data stream mining \cite{ABifet2010MA}. The source codes of the FDCM\_SSR and FDPC algorithms are provided by their authors.

\subsection{Experimental Settings}
Four benchmark datasets are used in the evaluation \cite{Dua2017UCI, Souza2015DS}. The data objects are randomly shuffled to alleviate any potential effects caused by the sorted streaming order. The drift interval of the Keystroke dataset is 200. The drift intervals of the other datasets are unknown. The Keystroke dataset is composed of 10 feature variables. The Network Intrusion dataset was collected from seven weeks of network traffic simulated in a military network environment. The Network Intrusion requests are divided into two classes: normal and malicious. All the text attributes of Network Intrusion are manually converted into enumerated values and represented by digits. The Forest Cover dataset consists of 54 cartographic variables. The last variables of the datasets represent the ground-truth labels of the corresponding data objects. The Columbia University Image Library (COIL)-100 dataset contains 7200 images of data objects belonging to 100 categories. The statistics of the datasets are summarized in Table \ref{tb:datasets}.

The HSRC algorithm is evaluated based on its clustering quality and temporal efficiency. Clustering quality is assessed via clustering purity and the $F$-measure \cite{Manning2008IR}. All the parameters of the competing algorithms are manually tuned to obtain their optimal results. In Algorithms \ref{alg:HSRC1}-\ref{alg:HSRC4}, HSRC involves three parameters: $\lambda$, $m$ and $\sigma $. Empirically speaking, parameter $\lambda$ should be relatively large if the input data streams are slightly contaminated by noise, and vice versa. Parameter $m$ controls the number of microclusters and is typically set as a multiple of the maximum number of classes contained in the data streams. For our experiments, we use a modified parameter ${m'}$ instead of $m$, which ranges from 1 to 2. Microcluster merging is not performed if the maximum number of classes is less than or equal to 2 when ${m'}=1$. Parameter $\sigma $ is disregarded in cases where no outliers are detected. Additionally, HSRC disables the fine-tuning operation by setting $w= 0$. Further details concerning the parameters are provided in the experiments.

\begin{figure*}[!htbp]
\centering
\subfloat[Network Intrusion]{
\label{fig:performance:a} 
\includegraphics[width=6cm]{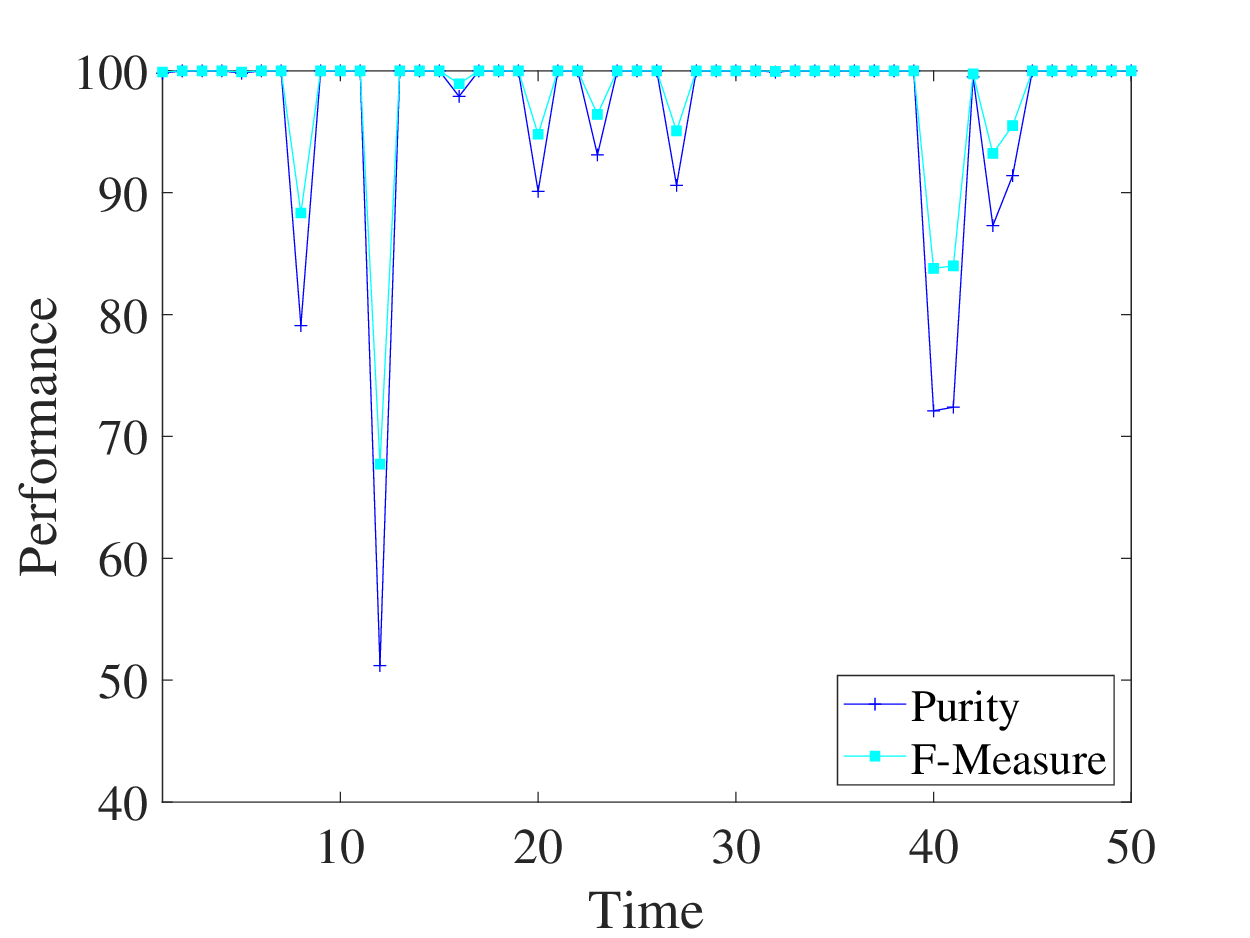}}
\subfloat[Forest Cover]{
\label{fig:performance:b} 
\includegraphics[width=6cm]{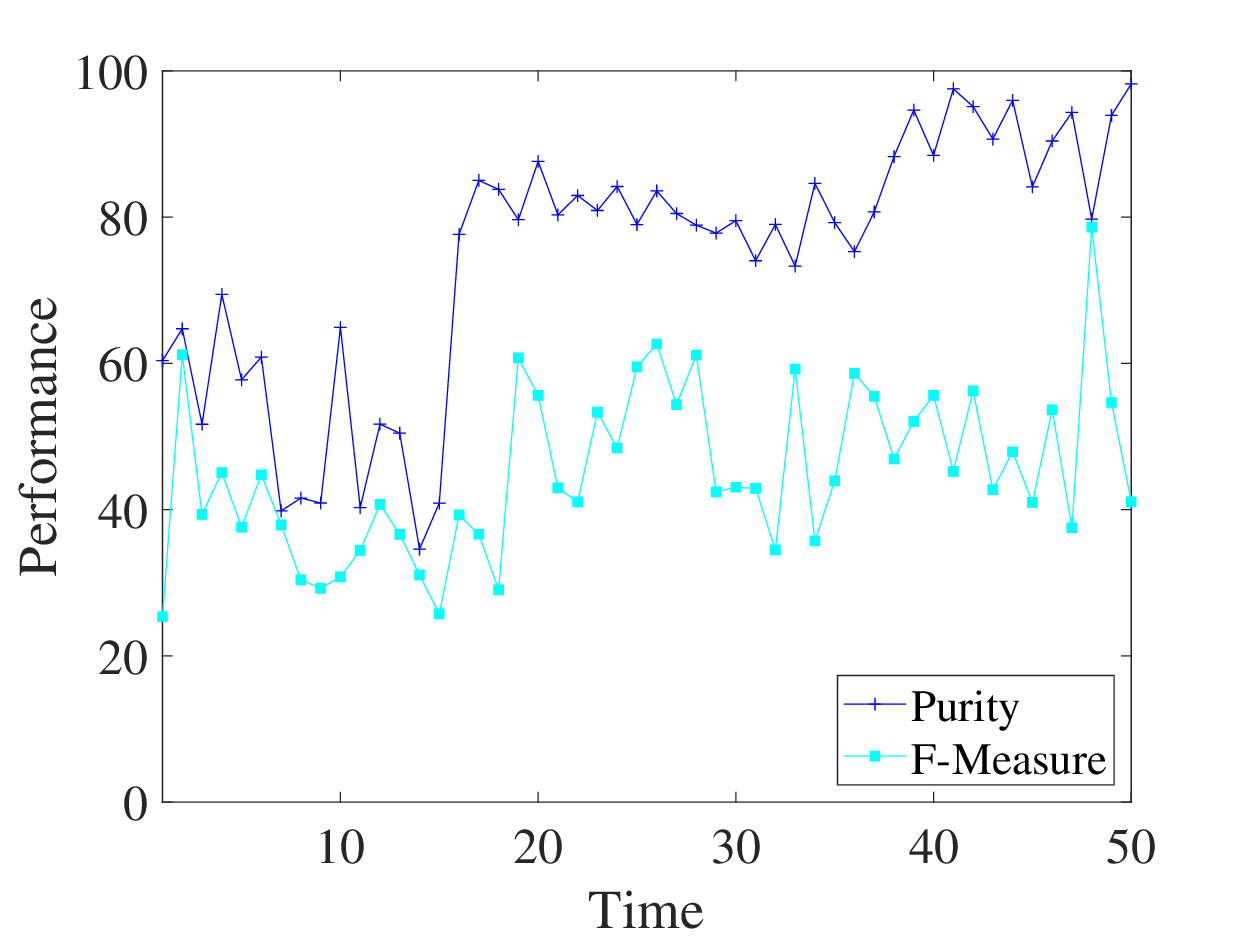}}
\caption{Online performance achieved on the first 50 windows of the Network Intrusion and Forest Cover datasets.}
\label{fig:performance} 
\end{figure*}

\subsection{Clustering Quality Evaluation}

\begin{table*}[!htbp]
\small
\setlength{\abovecaptionskip}{0pt}
\setlength{\belowcaptionskip}{0pt}
\setlength{\tabcolsep}{3pt}
\centering
\caption{Average clustering results (\%) produced over the entire stream of each dataset.}
\label{tb:nonstationary}
\begin{tabular}{c|ccccccccc|}
\hline
Datasets  & Metrics (\%) & HSRC & CluStream & ClusTree & CluStreamKM & StreamKM++ & FDCM\_SSR & FDPC \\
\hline
\multirow{2}{*}{Keystroke} &  Purity & \textbf{89.49} &  64 & 73 & 71  & 69 &  74.75 & \underline{80.06}\\
~ & $F$-measure & \textbf{89.03} & 74 & 75 & 72  & 70 &  \underline{75.27} & 70.59  \\
\hline
\multirow{2}{*}{Network Intrusion} & Purity & \textbf{96.48} & 93 & 91 & 94 & 92  &  94.33 & \underline{94.89} \\
~ &$F$-measure& \textbf{97.95} & 72 & 70 & 78  & 75 &  \underline{90.72} & 89.51 \\
\hline
\multirow{2}{*}{Forest Cover} & Purity & \textbf{74.58}  & 61 & 66 & 53 & 56 & 71.6 & \underline{71.88} \\
~ & $F$-measure & \textbf{45.28}  & 37 & 33 & 36 & 38 &   \underline{44.61} & 44.18 \\
\hline
\multirow{2}{*}{COIL-100}  & Purity & \textbf{67.74} & 41 & 39 & 44  & 43 &  \underline{51.1} & 44.96 \\
~ & $F$-measure & \textbf{64.24} & 40 & 42 & 38  & 39 &  \underline{47.97} & 26.87 \\
\hline
\end{tabular}
\end{table*}

\begin{table*}[!htbp]
\small
\setlength{\abovecaptionskip}{0pt}
\setlength{\belowcaptionskip}{0pt}
\setlength{\tabcolsep}{3pt}
\centering
\caption{The average computational costs required for a landmark window (seconds) in each dataset.}
\label{tb:cost}
\begin{tabular}{c|ccccccccc|}
\hline
Datasets  & HSRC & CluStream & ClusTree & CluStreamKM & StreamKM++  & FDCM\_SSR & FDPC \\
\hline
Keystroke & 0.49 & 0.15  &  \underline{0.9} & \textbf{0.09}  & 0.1 & 0.26 &  0.11 \\
\hline
Network Intrusion & 1.25 & 0.47 & 2.7 &  \textbf{0.24}  & 0.44 & 5.24 &   \underline{0.33} \\
\hline
Forest Cover & 1.22 &  \underline{0.29}  & 0.72 &  \textbf{0.27}  & 0.48 & 3.75  & 0.3 \\
\hline
COIL-100 & 5.73 & 1.71  & 2.21 & \textbf{1.55}  & 2.45 & 6.97 &  \underline{1.64} \\
\hline
\end{tabular}
\end{table*}

We evaluate the proposed algorithm on the streaming datasets. The four groups of HSRC parameters used for this experiment are (1) $\lambda=100$, $m'=1$, $w = 0$,  (2) $\lambda=0.1$, $m'=1$, $w = 1$, (3) $\lambda=12$, $m'=1$, $w = 0$ and (4) $\lambda=20$, $m'=1$, $w = 1$ . Compared with the other algorithms, the HSRC algorithm almost consistently obtains the best results in terms of clustering purity and the $F$-measure in the experiments. This confirms that our proposed method is very effective for high-dimensional data stream clustering cases with varying numbers of clusters. For example, HSRC achieves a high clustering purity level of 89.49\% on the Keystroke dataset and improves the clustering purity by at least 9.43\% over those of the other algorithms. We observe the same advantages in our proposed method, which contains more data object features. For example, HSRC achieves high clustering purities of 96.48\%, 74.58\% and 67.74\% for the remaining three datasets. Table \ref{tb:nonstationary} shows that it still significantly outperforms the other four algorithms in terms of the $F$-measure. These clustering results confirm that the relationships calculated from the sparse representations significantly improve the attained clustering performance, especially when the data objects derived from the data streams contain more features. In addition, density-based algorithms can identify clusters with arbitrary shapes in data streams and often perform well when the input data objects have only a few attributes. However, the clustering performance of these algorithms decreases as the number of data object attributes gradually increases.

We evaluate the online performance of the proposed HSRC method on the first 50 windows of two representative datasets, i.e., the Network Intrusion and Forest Cover. Fig. \ref{fig:performance} illustrates the online performance attained by the proposed HSRC method on these datasets. We can observe that HSRC almost achieves relatively stable clustering purity and $F$-measures on both datasets. For example, Fig. \ref{fig:performance:a} shows that the online clustering purity and $F$-measure values exceed  90\% for most of the landmark windows. Similarly, Fig. \ref{fig:performance:b} indicates that both metrics yielded by HSRC are above 25\%. These experimental results demonstrate the excellent stability of the online performance exhibited by the proposed HSRC method.

The time costs of all the competing algorithms are presented in Table \ref{tb:cost}. The density-based algorithms generally have lower computational costs than the other methods do. For example, CluStreamKM has a lower computational cost than the other algorithms do. However, HSRC has relatively reasonable computational costs in the experiments. This is because HSRC spends most of its time computing the affinity matrix using sparse representation.

Compared with $k$-means-based data stream clustering methods such as CluStreamKM and StreamKM++, HSRC employs a spectral clustering technique that also incorporates $k$-means to create microclusters. CluStreamKM and StreamKM++ compute the centroid of each CF vector via the original data objects. In contrast, HSRC uses an $l_1$-norm technique to construct an affinity matrix, which encodes the membership attributed of the high-dimensional data objects contained in the input data streams. HSRC captures the intrinsic structures of high-dimensional data objects. The experimental results show that HSRC significantly enhances the clustering purity and  $F$-measure values achieved for high-dimensional data streams.

\subsection{Robustness to Noise and Outliers}
We evaluate the robustness of these algorithms on a more challenging set of high-dimensional data streams. Four artificial pixel noise levels (5\%, 10\%, 15\%, and 20\%) are integrated into two datasets: (1) Network Intrusion and (2) Forest Cover. The locations of the corrupted data object attributes are chosen randomly, and the value of each selected location is replaced by a random number within the range $\left[ {0,1} \right]$.

\begin{table*}[!htbp]
\small
\setlength{\abovecaptionskip}{0pt}
\setlength{\belowcaptionskip}{0pt}
\setlength{\tabcolsep}{5pt}
\centering
\caption{Average clustering results (\%) produced on the Network Intrusion and Forest Cover datasets.}
\label{tb:noise}
\begin{tabular}{c|ccccccc|}
\hline
Datasets & Ratio (\%) & Metrics & HSRC & StreamKM++ & FDCM\_SSR & FDPC \\
\hline
\multirow{8}{*}{Network Intrusion} & \multirow{2}{*}{5\%} & Purity & \textbf{95.04} & 90 & 93.17 & \underline{93.9}	 \\
~ & ~ &$F$-measure & \textbf{96.93}  & 74  & \underline{89.53} & 88.79 \\
\cline{2-7}
~ & \multirow{2}{*}{10\%} & Purity & \textbf{94.95}  & 88  & 92.34 &  \underline{93.54} \\
~ & ~ & $F$-measure  & \textbf{96.45} & 72  & \underline{88.05}  & 80.85 \\
\cline{2-7}
~ & \multirow{2}{*}{15\%} & Purity & \textbf{94.4} & 86  &  91.68 &  \underline{92.15} \\
~ & ~ & $F$-measure  & \textbf{96.2} & 71  &  \underline{86.78} & 64.84 \\
\cline{2-7}
~ & \multirow{2}{*}{20\%} & Purity & \textbf{94.44} & 86  & \underline{91.54} & 90.04 \\
~ & ~ & $F$-measure & \textbf{96.4} & 70  & 85.88  & 56.23 \\
\hline
\multirow{8}{*}{Forest Cover} & \multirow{2}{*}{5\%} & Purity & \textbf{74.69} & 54 & 68.58  & \underline{71.03} \\
~ & ~ & $F$-measure & \textbf{43.61}  & 35 & \underline{41.14}  & 37.96 \\
\cline{2-7}
~ & \multirow{2}{*}{10\%} & Purity & \textbf{73.18} & 53 & 66.24  & \underline{68.81} \\
~ & ~ & $F$-measure & \textbf{39.27}  & 34 & \underline{38.16}  & 30.48\\
\cline{2-7}
~ & \multirow{2}{*}{15\%} & Purity & \textbf{71.96}  & 51 & 65.35  & \underline{66.14} \\
~ & ~ & $F$-measure & \textbf{38.65}  & 32 & \underline{33.5}  & 33.35 \\
\cline{2-7}
~ & \multirow{2}{*}{20\%} & Purity & \textbf{69.96}  & 50 & \underline{65.52}  & 65.18 \\
~ & ~ & $F$-measure & \textbf{37.68} & 29 & \underline{33.71}  & 30.88 \\
\hline
\end{tabular}
\end{table*}

Table \ref{tb:noise} shows the average clustering results produced for the four different noise levels. As expected, the performance of the algorithms slowly decreases as the percentage of noise increases. HSRC achieves consistently better clustering results than those of the other methods. For example, our method achieves average clustering purities of $95.04\%$, $94.95\%$, $94.4\%$ and $94.44\%$ on the Network Intrusion dataset. Regarding the $F$-measure, HSRC significantly outperforms the competing methods across the different noise levels, which indicates that the numbers of macroclusters are much closer to the ground-truth numbers of clusters contained in the landmark windows. However, the clustering performance of the other competing algorithms dramatically decreases as the percentage of noise slowly increases. We also observe that HSRC still retains the same advantages on the Forest Cover dataset. For example, the clustering purity and  $F$-measure of HSRC are $74.69\%$ and $43.61\%$, respectively under a noise percentage of 5\%. HSRC consistently achieves better clustering performance than that of the competing approaches under the other three noise percentages (10\%, 15\%, and 20\% ). On the one hand, this highlights the benefit of estimating the relationships among data objects via sparse representation. On the other hand, this finding demonstrates that the ${l_{2,1}}$-norm effectively characterizes the noise term in Problem \eqref{eq:sr4}. These experimental results demonstrate the robustness of the proposed HSRC method under noisy circumstances.

The proposed outlier detection mechanism is evaluated on the Forest Cover dataset. We select the first 20 landmark windows, and each window contains 1,000 data objects. In each trial, two successive landmark windows are employed, and the experiment is repeated 10 times. Half of the data objects contained within these windows are segmented into clusters, whereas the remaining data objects are treated as potential outliers for testing purposes. The effectiveness of the outlier detection mechanism is then verified via the SRV from Eq. \eqref{eq:srv}. For comparison, we employ the 1-NN method, which is a general outlier detection technique that calculates the Euclidean distance between a test data object and its nearest neighbor within a landmark window. The outlier detection errors are classified into two categories: valid data objects misidentified as outliers and actual outliers incorrectly classified as valid data. The average outlier error rate for HSRC is 4.21\%, whereas it is 6.64\% for 1-NN across the 10 trials. Compared with 1-NN, HSRC reduces the average outlier error rate by 2.43\%.

\subsection{Sensitivity Analysis}
In Algorithms \ref{alg:HSRC1}-\ref{alg:HSRC4}, HSRC has two main parameters: $\lambda$ and $m$. We examine the sensitivity of different combinations of parameters $\lambda$ and $m$. In particular, we set parameter $w = 1$ for the Network Fusion and Forest Cover datasets. Empirically, parameter $\lambda$ is closely related to the noise level prior of the data stream. Hence, we usually set relatively large values for parameter $\lambda$ in HSRC if the given data stream is slightly contaminated by noise. In addition, parameter $m$ represents the number of microclusters. For convenience, parameter $m'$ represents a multiple of the maximum number of classes contained in a data stream. Parameter $m'$  ranges from 1 to 2 in the experiments.

\begin{figure}[!htbp]
\centering
\subfloat[$m'=1$]{
\label{fig:network:a} 
\includegraphics[width=4.2cm]{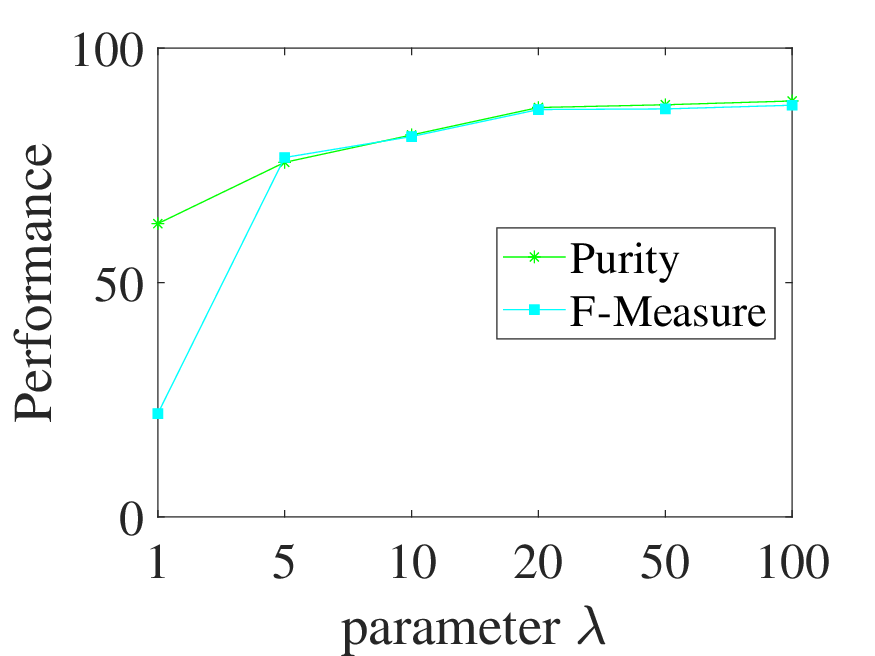}}
\subfloat[$m'=2$]{
\label{fig:network:b} 
\includegraphics[width=4.2cm]{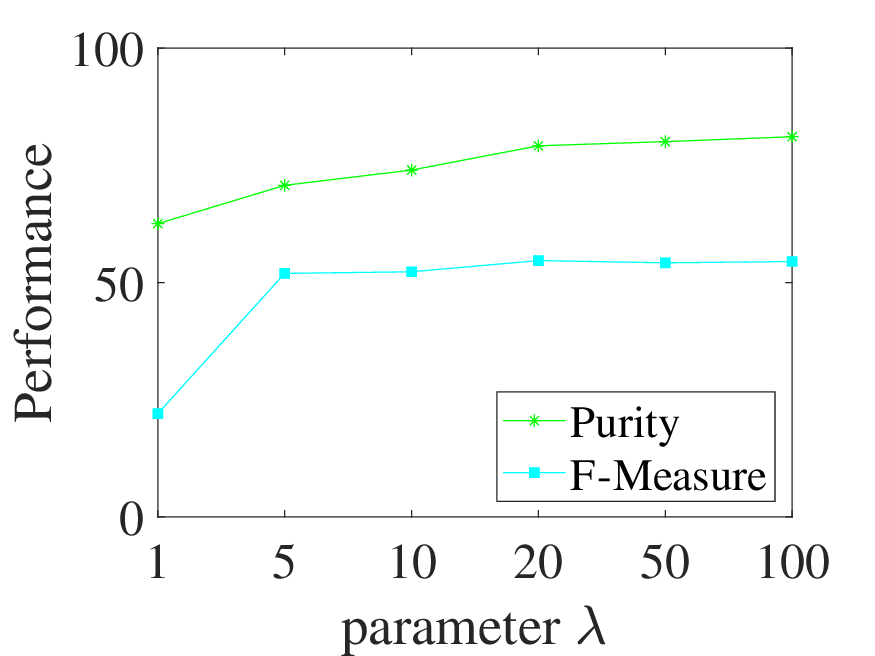}}
\caption{Clustering results obtained with different values of $\lambda$ when using the first 50 windows of the Network Fusion dataset.}
\label{fig:network} 
\end{figure}

\begin{figure}[!htbp]
\centering
\subfloat[$m'=1$]{
\label{fig:forest:a} 
\includegraphics[width=4.2cm]{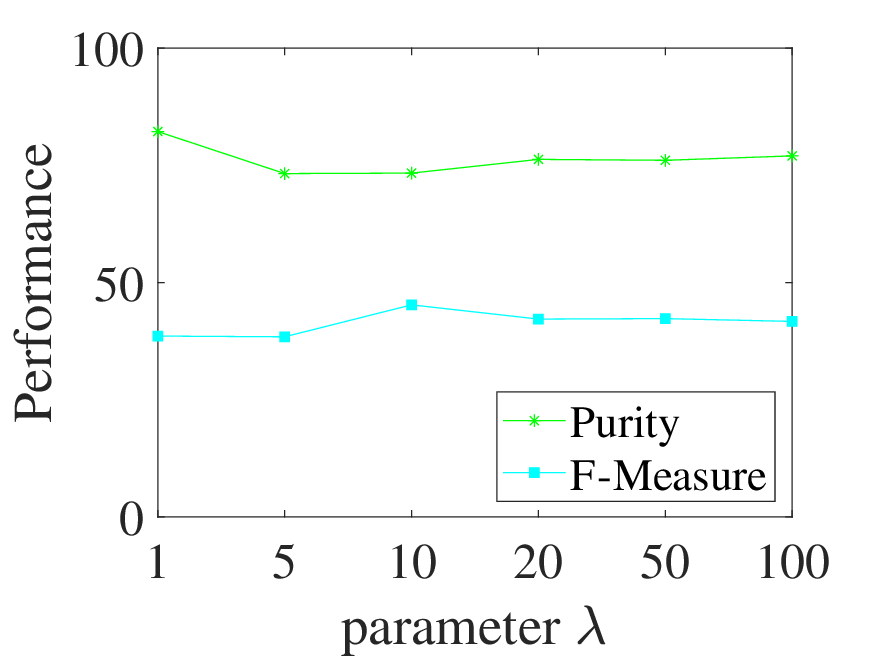}}
\subfloat[$m'=2$]{
\label{fig:forest:b} 
\includegraphics[width=4.2cm]{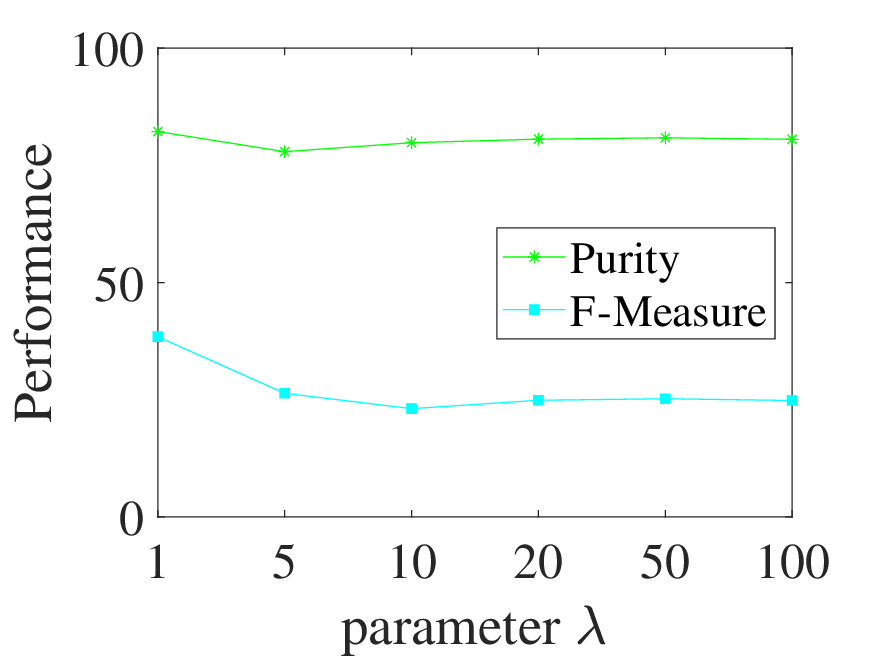}}
\caption{Clustering results obtained with different values of $\lambda$ when  using the first 50 windows of the Forest Cover dataset.}
\label{fig:forest} 
\end{figure}

Figs. \ref{fig:network} and \ref{fig:forest} show the clustering performance achieved on the Network Fusion and Forest Cover datasets, respectively, in terms of the clustering purity and $F$-measure values produced with different combinations of parameters $\lambda$ and $m'$. Specifically, Fig. \ref{fig:network} indicates that HSRC performs well across a large range of $\lambda$ values when $m'=1$. However, the $F$-measure declines significantly as $m'$ increases from 1 to 2. This is because the number of clusters rapidly decreases as the number of macroclusters gradually increases during the experiments. Similar cases can also be observed in Fig.  \ref{fig:forest}. For example, the $F$-measure dramatically decreases when $m'$ gradually increases from 1 to 2 in Fig. \ref{fig:forest}. Therefore, relatively large ranges of  $\lambda$ yield satisfactory clustering results when $m'=1$.

\section{Conclusion}
\label{sec:Conclusion}
In this paper, we propose an HSRC algorithm for clustering high-dimensional data streams. Unlike the existing clustering techniques that rely on Euclidean distance computations, HSRC utilizes an   norm technique to capture the intrinsic structures of high-dimensional data objects. It automatically selects the appropriate number of neighboring data objects. This ensures that the highly correlated data objects of clusters are grouped together. HSRC merges microclusters into macroclusters by introducing the SSD. Moreover, fine-tuning is employed to refine the macroclusters. Data object representatives are selected from each macrocluster by sorting the SRVs of the data objects. These representatives are then employed as dictionary samples for the next landmark window. The proposed HSRC method significantly enhances the relationships among data objects, which improves the generalization of sparse representation in high-dimensional data stream clustering tasks. Furthermore, HSRC effectively estimates outlier candidates based on their SRVs. Our extensive experiments conducted on high-dimensional datasets demonstrate the superiority of the proposed HSRC method over several state-of-the-art approaches.

 \bibliographystyle{IEEEtran}
\bibliography{hsrc}


\vspace{-1cm}
\begin{IEEEbiography}[{\includegraphics[width=1in,height=1.25in,clip,keepaspectratio]{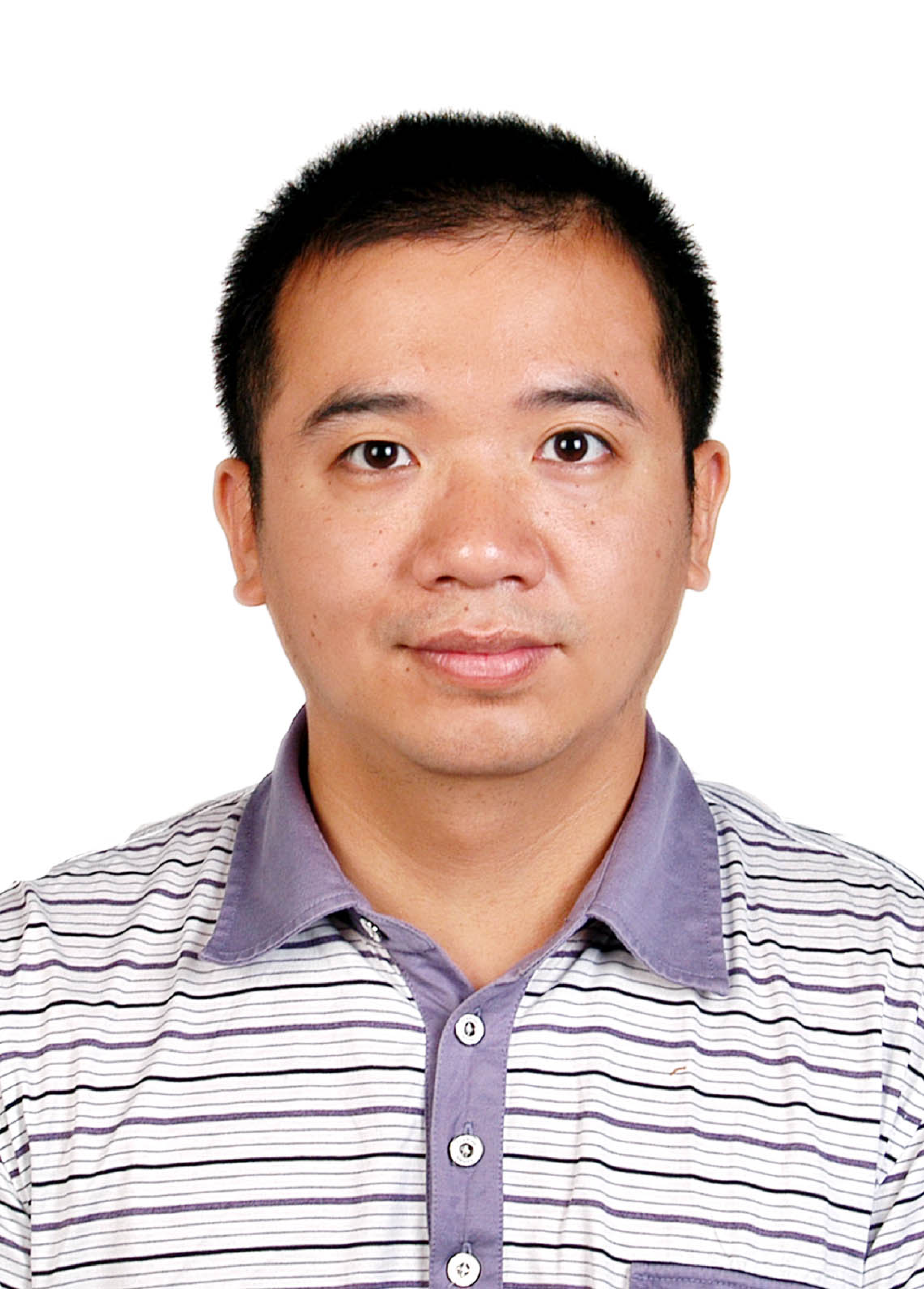}}]%
{Jie Chen} received the BSc degree in Software Engineering, MSc degree and PhD degree in Computer Science from Sichuan University, Chengdu, China, in 2005, 2008 and 2014, respectively. He is currently an Associate Professor in the College of Computer Science, Sichuan University, China. His current research interests include machine learning, big data analysis and deep neural networks.
\end{IEEEbiography}

\vspace{-1cm}
\begin{IEEEbiography}[{\includegraphics[width=1in,height=1.25in,clip,keepaspectratio]{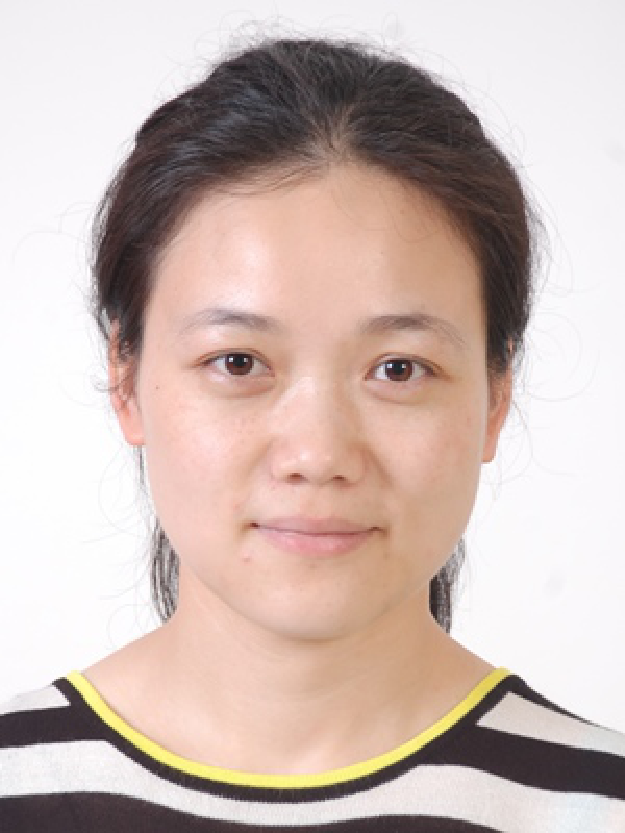}}]%
{Hua Mao} received the B.S. degree and M.S. degree in Computer Science from University of Electronic Science and Technology of China (UESTC) in 2006 and 2009, respectively. She received her Ph.D. degree in Computer Science and Engineering from Aalborg University, Denmark in 2013. She is currently a Senior Lecturer in Department of Computer and Information Sciences, Northumbria University, U.K. Her current research interests include Deep Neural Networks and Big Data.
\end{IEEEbiography}

\vspace{-1cm}
\begin{IEEEbiography}[{\includegraphics[width=1in,height=1.25in,clip,keepaspectratio]{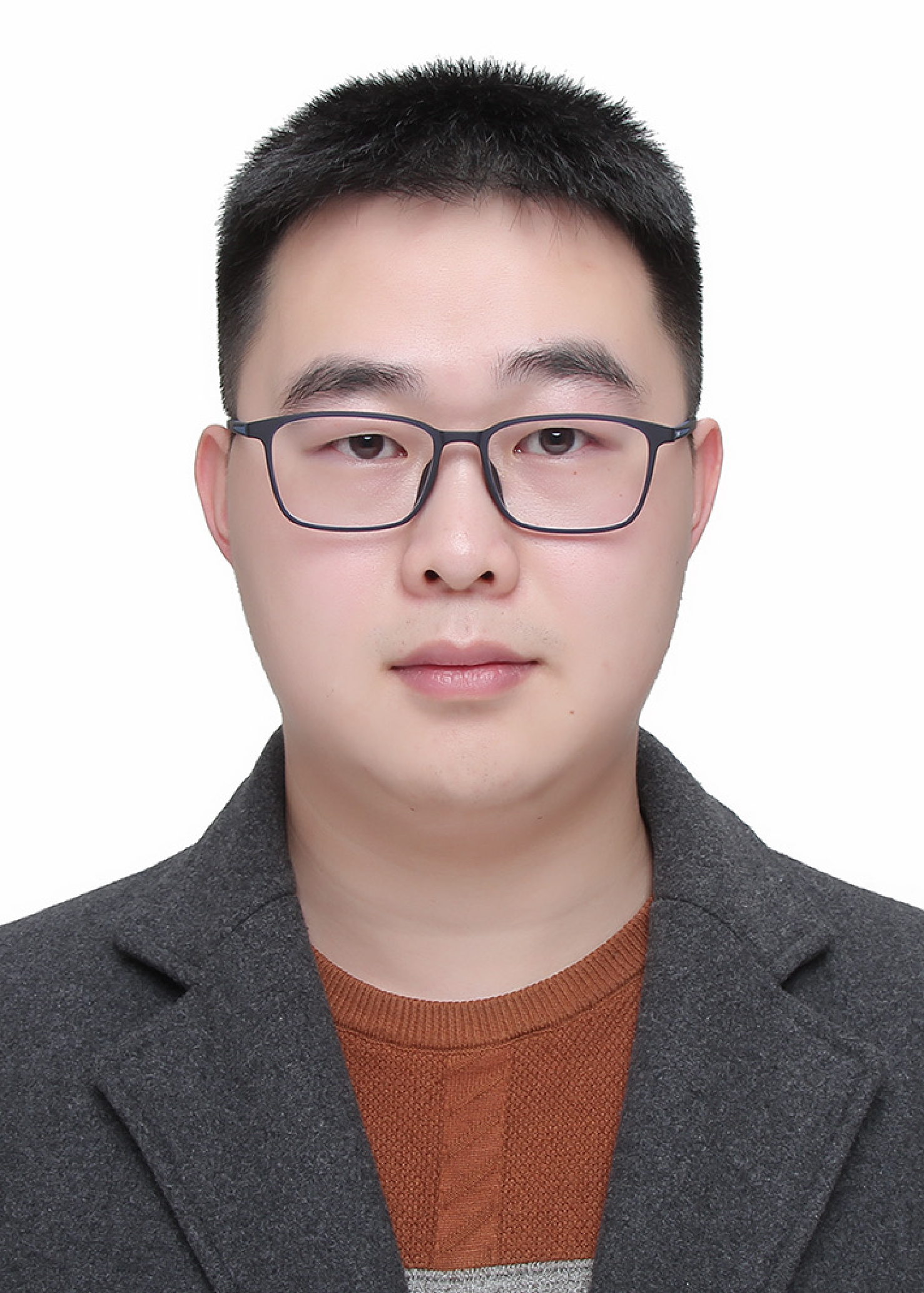}}]%
{Yuanbiao Gou} received the B.E. degree from the School of Software Engineering, Sichuan University, Chengdu, China, in 2019. He is currently pursuing the Ph.D. degree with the School of Computer Science, Sichuan University, under the supervision of Prof. Xi Peng. His current research interests include image processing and computer vision.
\end{IEEEbiography}

\vspace{-1cm}
\begin{IEEEbiography}[{\includegraphics[width=1in,height=1.25in,clip,keepaspectratio]{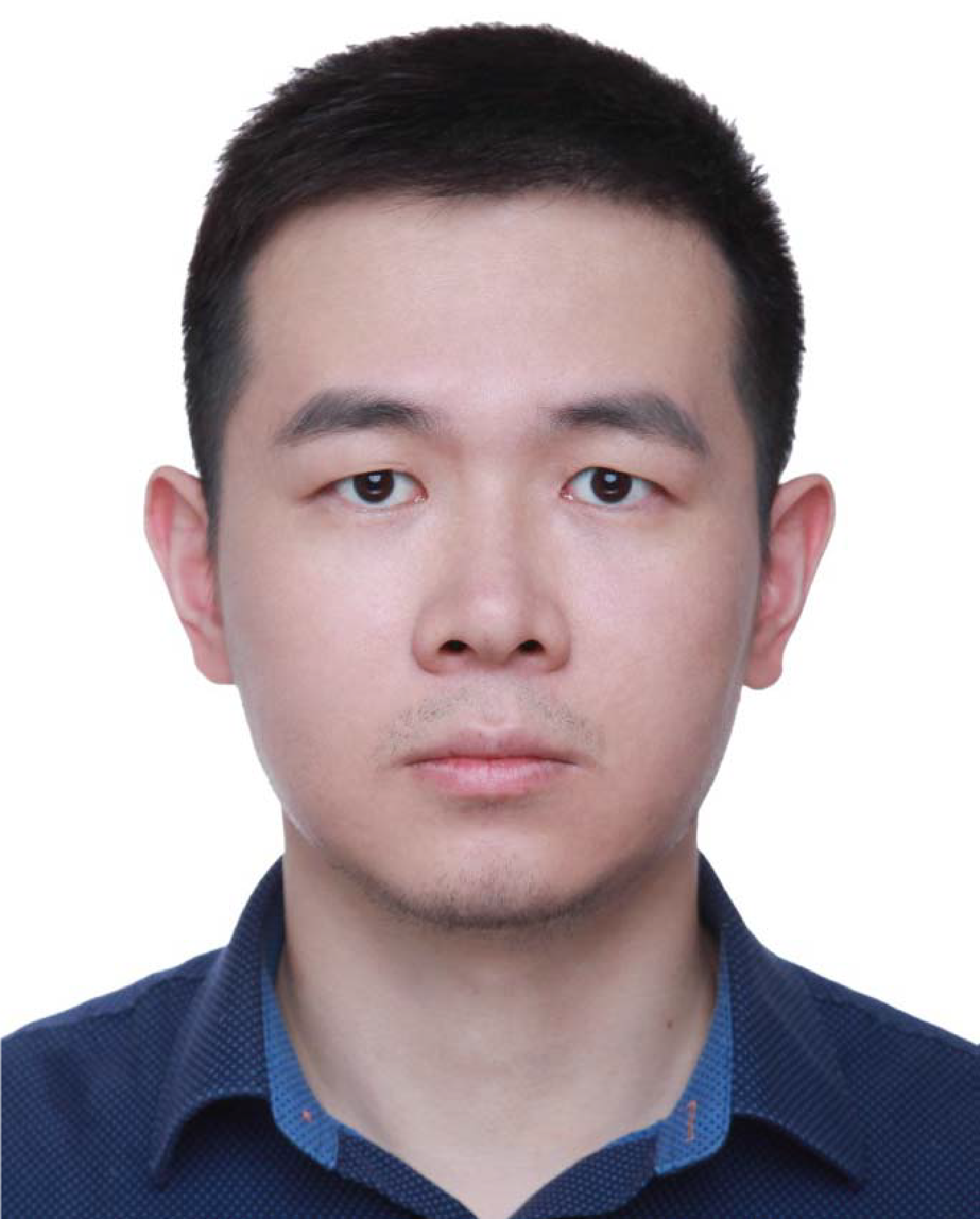}}]%
{Xi Peng} (Member, IEEE) is currently a Full Professor with the College of Computer Science, Sichuan University. His current research interest includes machine intelligence and has authored more than 50 articles in these areas. He has served as an Associate Editor/Guest Editor for six journals, including the {\em IEEE Transactions on SMC: Systems} and {\em IEEE Transactions on Neural Network And Learning Systems} and the Area Chair/Senior Program Committee Member for the conferences such as IJCAI, AAAI, and ICME.
\end{IEEEbiography}

\vfill

\end{document}